\documentclass[journal]{IEEEtran}

\usepackage{graphicx}
\usepackage{xcolor}
\usepackage{amsmath,amssymb,amsfonts}
\usepackage{amsthm}
\usepackage{mathrsfs}
\usepackage{soul}
\usepackage{booktabs}
\usepackage{array}
\usepackage{tabularx}
\usepackage{multirow}
\usepackage{longtable}
\usepackage{siunitx}
\usepackage{tikz}

\usepackage[ruled,vlined]{algorithm2e}

\begin{document}


\title{Sim-to-Real Transfer in Deep Reinforcement Learning for Bipedal Locomotion}

\author{Lingfan Bao, Tianhu Peng, and Chengxu Zhou
\thanks{Lingfan Bao, Tianhu Peng, and Chengxu Zhou are with the Department of Computer Science,
University College London, London WC1E 6BT, United Kingdom (e-mail: chengxu.zhou@ucl.ac.uk).}}
\maketitle

\begin{abstract}
This chapter addresses the critical challenge of simulation-to-reality (sim-to-real) transfer for deep reinforcement learning (DRL) in bipedal locomotion. After contextualizing the problem within various control architectures, we dissect the ``curse of simulation'' by analyzing the primary sources of sim-to-real gap: robot dynamics, contact modeling, state estimation, and numerical solvers. Building on this diagnosis, we structure the solutions around two complementary philosophies. The first is to shrink the gap through model-centric strategies that systematically improve the simulator's physical fidelity. The second is to harden the policy, a complementary approach that uses in-simulation robustness training and post-deployment adaptation to make the policy inherently resilient to model inaccuracies. The chapter concludes by synthesizing these philosophies into a strategic framework, providing a clear roadmap for developing and evaluating robust sim-to-real solutions.

\end{abstract}


\section{Introduction}
\label{sec:introduction}

Bipedal robots, machines that walk on two legs, are compelling platforms for operation in human-centric and natural environments. They can climb stairs, step over irregular obstacles, traverse narrow passages, and access spaces that are impractical for wheeled platforms. Their anthropomorphic form factor also enables natural interaction with tools and infrastructure designed for humans, making them suitable for disaster response, healthcare, logistics, and industrial applications.

\begin{figure}
\centering
\includegraphics[trim={0cm 0.2cm 0cm 0.5cm},clip,width=\linewidth]{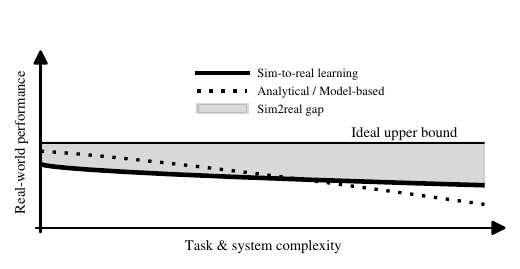}
\caption{Comparison of model-based control and sim-to-real learning in terms of real-world performance. As task and system complexity increase, performance declines; however, sim-to-real learning (solid) maintains higher levels than analytical or model-based methods (dotted). The shaded region indicates the sim-to-real gap relative to the ideal upper bound determined by hardware, sensing, and safety constraints.}
\label{fig:modelbased_vs_learningbased}
\end{figure}

Bipedal locomotion remains challenging because of its high dimensionality, underactuation, and intermittent contacts. Model-based methods struggle with complex dynamics, whereas deep reinforcement learning (DRL) has achieved impressive simulation results in bipedal locomotion through trial and error. As shown in Fig.~\ref{fig:modelbased_vs_learningbased}, DRL achieves more robust performance than model-based control, particularly as task complexity increases. Most controllers adopt either end-to-end policies that map observations to actions or hierarchical policies that decouple high-level (HL) intent from low-level (LL) execution. Both approaches perform well in simulation but transfer unreliably to hardware, a limitation known as the sim-to-real gap.

This transfer difficulty raises three framing questions: \textit{Why do we train in simulation before deployment? What precisely causes the sim-to-real gap? If this workflow is necessary, how can we improve it?} These questions, together with our control architectures, define the scope of this chapter.

To begin with the first question, learning locomotion behaviors requires millions of trials and inevitably explores unsafe or costly failure states on hardware. In contrast, simulation enables high-throughput parallel rollouts, precise control of conditions, and reproducibility, making data collection far more efficient and easier to diagnose. Consequently, the practical workflow is to train in a high-fidelity simulator and then transfer the resulting policy to the robot.

Addressing the sim-to-real challenge requires two complementary strategies. The first is to reduce the gap by improving simulator fidelity and calibrating unknown parameters. The second is to strengthen the policy so that it can tolerate residual discrepancies, providing robustness to environmental variations and the ability for online adaptation. In the remainder of this chapter, we organize these strategies into three levers: pre-training alignment, robustness training, and post-deployment adaptation.

This chapter provides a structured overview of sim-to-real transfer for bipedal locomotion with DRL. Section~\ref{sec:control_scheme} reviews control architectures, comparing end-to-end policies with hierarchical schemes. Section~\ref{sec:cause_sim2real} analyzes the primary sources of the simulation-to-reality (sim-to-real) gap. Section~\ref{sec:bridge-sim-real} presents bridging strategies in terms of three levers: pre-training alignment, robustness training, and post-deployment adaptation. Section~\ref{sec:discussion} discusses the limitations of current sim-to-real methods, distills lessons from theory to practice, and offers a future outlook. Section~\ref{sec:conclusion} concludes with practical guidelines and open research problems.


\section{Control Architectures}
\label{sec:control_scheme}

\begin{figure}[t]
\centering
\includegraphics[trim={2.5cm 0.4cm 2cm 0.7cm},clip,width=\linewidth]{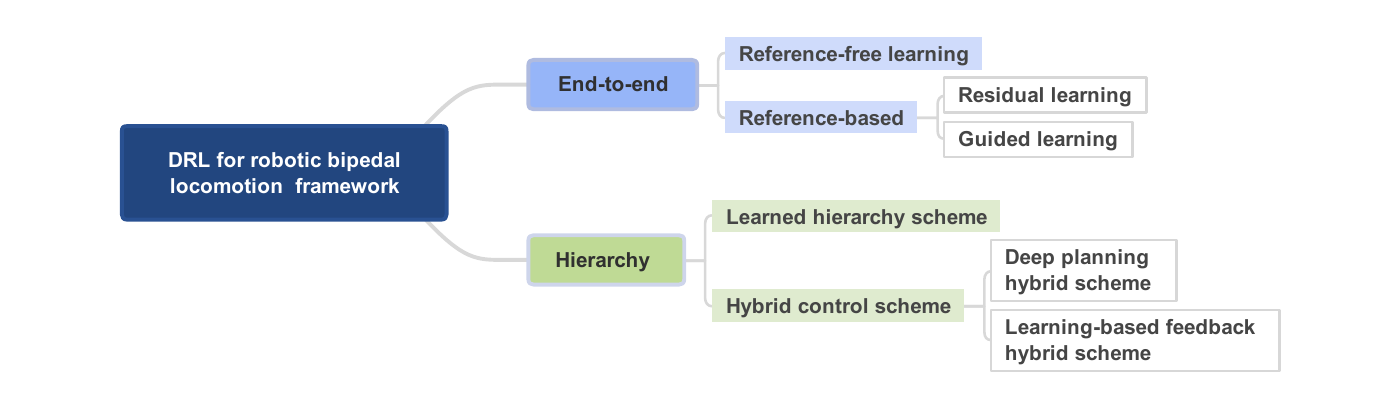}
\caption{\textbf{Classification of DRL-based control schemes.}}
\label{fig:framework_catalog}
\end{figure}

Sim-to-real DRL for bipedal locomotion is commonly organized into two control architectures: end-to-end and hierarchical. End-to-end policies map observations directly to low-level actions (for example, torques or impedance targets), allowing the robot to discover viable behaviors without manually specified intermediate structures. Hierarchical architectures, in contrast, decompose decision making into multiple layers. The two categories are summarized in Fig.~\ref{fig:framework_catalog}. The choice between them represents a trade-off between simplicity and expressive capacity on one side, and interpretability, explicit constraint handling, and composability on the other, with significant implications for sim-to-real transfer.

The remainder of this section contrasts these approaches and their policy representations, highlights assumptions and limitations, and provides practical guidance for sim-to-real deployment.

\subsection{End-to-End Framework}
\label{sec:e2e}

Formally, an end-to-end DRL policy $\pi:\mathcal{X}!\rightarrow!\mathcal{U}$ maps sensory inputs $\mathcal{X}$, such as images, lidar, or proprioceptive feedback \cite{2017_compare_actionspace_RL}, together with user commands~\cite{2021_siekmann_sim2real_nonreference_perodicreward_DRL_e2e_LSTM_PPO_cassie} or predefined references~\cite{2021_UCB_hybridrobotics_sim2real_referencebased_HZD_gaitlibrary_e2epolicy_drl_Cassie_lowpassfilter}, to joint-level actions $\mathcal{U}$. Here, $\mathcal{X}$ denotes the observation space, $\mathcal{U}$ the action space, and $\pi(\cdot)$ the policy. In practice, lightweight state estimation and safety filters are typically retained.

\begin{figure*}[ht]
\centering
\includegraphics[trim={2.5cm 9.2cm 2.8cm 9.2cm},clip,width=\textwidth]{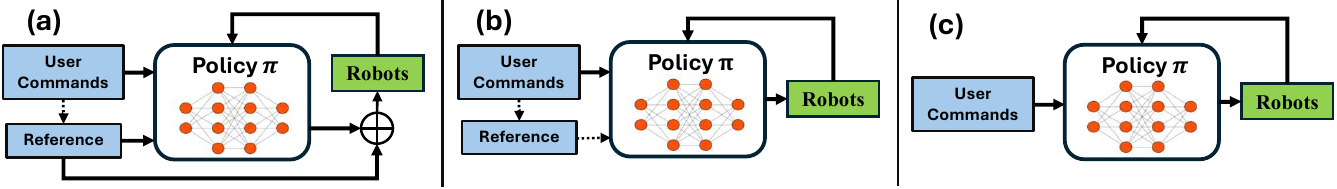}
\caption{End-to-end locomotion policy variants. (a) Residual: the policy adds a bounded correction to a nominal controller. (b) Guided: the policy is conditioned on a motion or controller reference. (c) Reference-free: the policy learns solely from task rewards. Feedback arrows indicate onboard estimates returned to the policy.}
\label{fig:e2e_framework}
\end{figure*}

End-to-end methods can be categorized by the amount of prior structure they employ. Reference-based policies condition on motion templates or commanded trajectories, which improves sample efficiency and targeting but introduces biases and limits feasibility. Residual or guided variants can help mitigate these drawbacks. Reference-free policies, by contrast, learn purely from task rewards, providing greater expressiveness but relying heavily on reward shaping, exploration, and robustness for successful transfer. Figure~\ref{fig:e2e_framework} summarizes three representative training strategies.

\paragraph{Reference-Based Learning}

Reference-based learning leverages prior knowledge generated offline through methods such as trajectory optimization (TO) or motion capture systems. This predefined reference typically contains data related to the robot’s joint movements or preplanned trajectories, providing a foundation for the policy to develop locomotion skills by following established motion patterns. In general, this approach can be divided into two primary methods: (i) residual learning and (ii) guided learning.

The residual learning framework employs a policy that adjusts motor commands by applying action offsets relative to the current reference joint positions. This enables the bipedal robot to achieve dynamic locomotion through error compensation. The state space includes proprioceptive information such as trunk position, orientation, velocity, angular velocity, joint angles, and joint velocities, which together provide the necessary sensory data for real-time adjustments. Actions are defined by offsets $\delta a$, representing deviations from the desired joint positions $\hat{a}$, and the final motor commands are given by $a = \hat{a} + \delta a$. The reward function encourages the policy to optimize locomotion performance by considering (a) how closely the robot’s active joint angles match the reference angles, (b) how effectively the robot responds to user commands, and (c) additional terms that enhance stability and smoothness of motion. This formulation allows the robot to adapt to dynamic conditions while maintaining balance and control.

Guided learning, in contrast, trains the policy to directly output joint-level commands as actions $a$ without adding a residual term. The state space is the same as that used in residual learning. In this approach, the reward structure focuses on accurately imitating predefined reference trajectories, ensuring close alignment between the policy outputs and the reference motion.

\paragraph{Reference-Free Learning}

In reference-free learning, the policy is trained using a carefully designed reward function rather than predefined trajectories. This approach allows the policy to explore a wider range of gait patterns and adapt to unpredictable terrains, thereby enhancing flexibility and innovation in the learning process. The action and observation spaces are similar to those in the guided-learning framework; however, the reward structure differs substantially. Instead of emphasizing imitation of reference motions, the reward focuses on developing efficient gait patterns that capture the distinctive characteristics of bipedal locomotion~\cite{2023_vanmarum_visionDRL_studentteacher_irregularterrain_PPO_periodicrewardfunction}.

The concept of reference-free learning was initially explored in simulated physics environments using simplified bipedal models. A pioneering framework demonstrated that symmetric gaits could be learned entirely from scratch without motion capture data~\cite{2018_wenhao_yu_DRL_withoutpredefine_symmetrygait_PPO_jointanglePD_}. This framework introduced a specialized term in the loss function and employed curriculum learning to gradually shape the gait. Another significant contribution involved a learning method that enabled a robot to traverse stepping stones through curriculum-based training on the Cassie platform, although this method has not yet been validated outside of simulation~\cite{2020_zhaoming_drl_steppingstones_PPOwithactorcritic_referencefree_simulation}.

This learning paradigm promotes the discovery of novel strategies and behaviors that may not emerge through imitation alone. However, the absence of reference guidance can make training costly, time-consuming, and occasionally infeasible for complex tasks. Furthermore, the success of reference-free learning depends heavily on the design of the reward function, which poses particular challenges for dynamic behaviors such as jumping.

\subsection{Hierarchical Framework}

\begin{figure*}
\centering
\includegraphics[trim={3.6cm 7.5cm 4cm 7.3cm},clip,width=\textwidth]{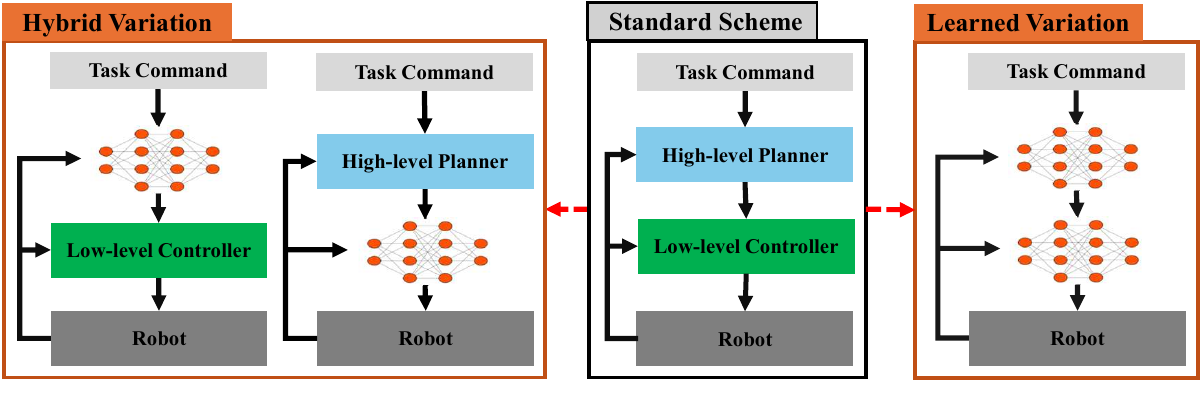}
\caption{Hierarchical control architectures for a bipedal robot. The central panel (``Standard Scheme'') presents the canonical hierarchy: a task command is processed by an HL planner, whose output is executed by an LL controller to actuate the robot. The left panel (``Hybrid Variation'') illustrates configurations in which one layer is learned while the other remains model-based (either a learned HL planner with a classical LL controller or the reverse). The right panel (``Learned Variation'') implements both HL planning and LL control as learned policies arranged in a two-layer hierarchy.}
\label{fig:Hierarchy_scheme_flow_chart}
\end{figure*}

Unlike end-to-end policies that directly map sensory inputs to motor commands, hierarchical control decomposes locomotion into multiple decision stages. A canonical architecture includes two modules: an HL planner and an LL controller. Either module can be model-based or learned, allowing task-specific customization. We categorize variants into three types, summarized in Fig.~\ref{fig:Hierarchy_scheme_flow_chart}. The HL planner issues abstract goals that the LL controller translates into executable trajectories and motor commands, while sensor feedback flows upward for online correction. Because the layers operate at different frequencies, communication and synchronization between them are central design considerations. We begin with deep-planning hybrids, in which a learned planner generates task-space setpoints and a model-based controller executes them at the joint level.

\paragraph{Deep-Planning Hybrid Scheme}

Several studies have demonstrated the integration of a learned HL planner with a model-based controller to achieve task-space objectives. One representative framework optimizes task-space performance without explicitly handling joint-level balance dynamics~\cite{2021_duan_DRL_task-spaceaction_hiarachycotnrolscheme_inversedynamiccontroller}. This system combines a residual-learning planner with an inverse dynamics controller, enabling precise translation of task-space commands into joint-level actions and improving velocity tracking, foot placement, and height control. Further advancements include hybrid frameworks that merge HZD-based residual deep planning with model-based regulators to correct trajectory errors, demonstrating robustness, training efficiency, and accurate velocity tracking~\cite{2022_DRL_cascade_motionplanningpolicy_HZD_torso_PPOwithRNN}. These approaches have successfully transferred from simulation to physical platforms such as Cassie.

\paragraph{Feedback-DRL Control Hybrid Scheme}

Unlike the end-to-end formulation in Section~\ref{sec:e2e}, this framework places a DRL policy at the LL while an HL planner provides terrain-aware footstep or center-of-mass (CoM) references and safety margins. Gait-library architectures follow this pattern~\cite{2021_Kevin_gaitlibrary_heirarchy_reducedorder_gaitlibrary_sim2real}, although static references restrict adaptability to varying terrain. A more flexible design combines a conventional footstep planner with an LL DRL controller that tracks planned steps and headings~\cite{2022_Singh_humanoid_e2erlwithfootplannar_hierarchyscheme_PPO_simulation_HRP-5P}, initially demonstrated in simulation. Later work using a similar HL–LL division achieved successful sim-to-real transfer, enabling omnidirectional walking, obstacle avoidance, and stair negotiation~\cite{2023_wangsong_DRL_footsteptrakcing_hybridscheme_experiment_PPO_stair_task_LIPM_curriculumlearning}. Injecting HL feedback, such as CoM or end-effector states, into the LL policy further enhances robustness.

In this scheme, robots are typically equipped with model-based feedback controllers or interpretable methods that handle basic locomotion skills such as walking. The addition of a learned HL layer focuses on strategic task-space goals, thereby extending locomotion capabilities and enabling advanced navigation behaviors.

\paragraph{Learned Hierarchical Scheme}

The learned hierarchy framework integrates a learned HL planner with a learned LL controller. Training generally begins with refining the LL policy to ensure balance and fundamental locomotion, followed by developing an HL policy that directs the robot toward specific targets. This layered process supports structured autonomy and facilitates scalable training.

Early hierarchical systems were validated in simulation~\cite{2017_Xuebinpeng_animation_deeploco_DRL_hiarachysystem_referencemotion_deepconvolutionalneuralnewtwork_jointangle_soccer.}. In these systems, the LL controller tracked motion references derived from human demonstrations or TO, while the HL planner issued long-horizon navigation goals, enabling tasks such as soccer dribbling. Subsequent work incorporated imitation learning to reproduce agile, human-like skills in physics simulation~\cite{2018_deepmimic_animation_PPO_deepmimic_desrieddirection_physics-based_animation_ATLAS_task}.

In these learned hierarchies, the HL layer governs overall strategy and goals and can be further refined through continued training. Recent studies report successful sim-to-real implementations on humanoid robots, including human-motion-conditioned whole-body control and outdoor locomotion~\cite{2024_fu_humanplus,2024_Radosavovic_SciRobotics_humanoid_RL}. However, training multiple layers remains computationally expensive and simulation-centric, which can cause overfitting and poor generalization~\cite{2023_Wei_learned_hierarchy_framework_wheeled_bipedalrobot}. Reliable deployment requires clean interfaces and robust feedback between layers.

While hierarchical control improves modularity and interpretability, it also introduces distinct challenges for sim-to-real transfer. The HL planner relies on assumptions about dynamics, terrain, and sensing that may not hold in reality. The LL controller must cope with contact, friction, latency, and actuator limits that differ between simulation and hardware. Moreover, the interface between levels imposes bandwidth and timing constraints that are easily maintained in simulation but fragile on real systems. These effects can compound, as small modeling errors in the planner propagate to aggressive setpoints that the controller executes with mismatched gains or delays.

\section{Crossing Simulation and Reality}
\label{sec:bridge-sim-real}

Building on the control schemes introduced in Section~\ref{sec:control_scheme}, we now address how policies trained in simulation can be effectively transferred to hardware. Most contemporary bipedal locomotion frameworks follow this train-in-simulation, deploy-on-hardware paradigm. Although transfer requires considerable effort, it remains the most practical and scalable approach for developing real-world controllers.

\begin{figure*}[t]
    \centering
    \includegraphics[trim={0cm 4.3cm 0cm 4.8cm},clip,width=\textwidth]{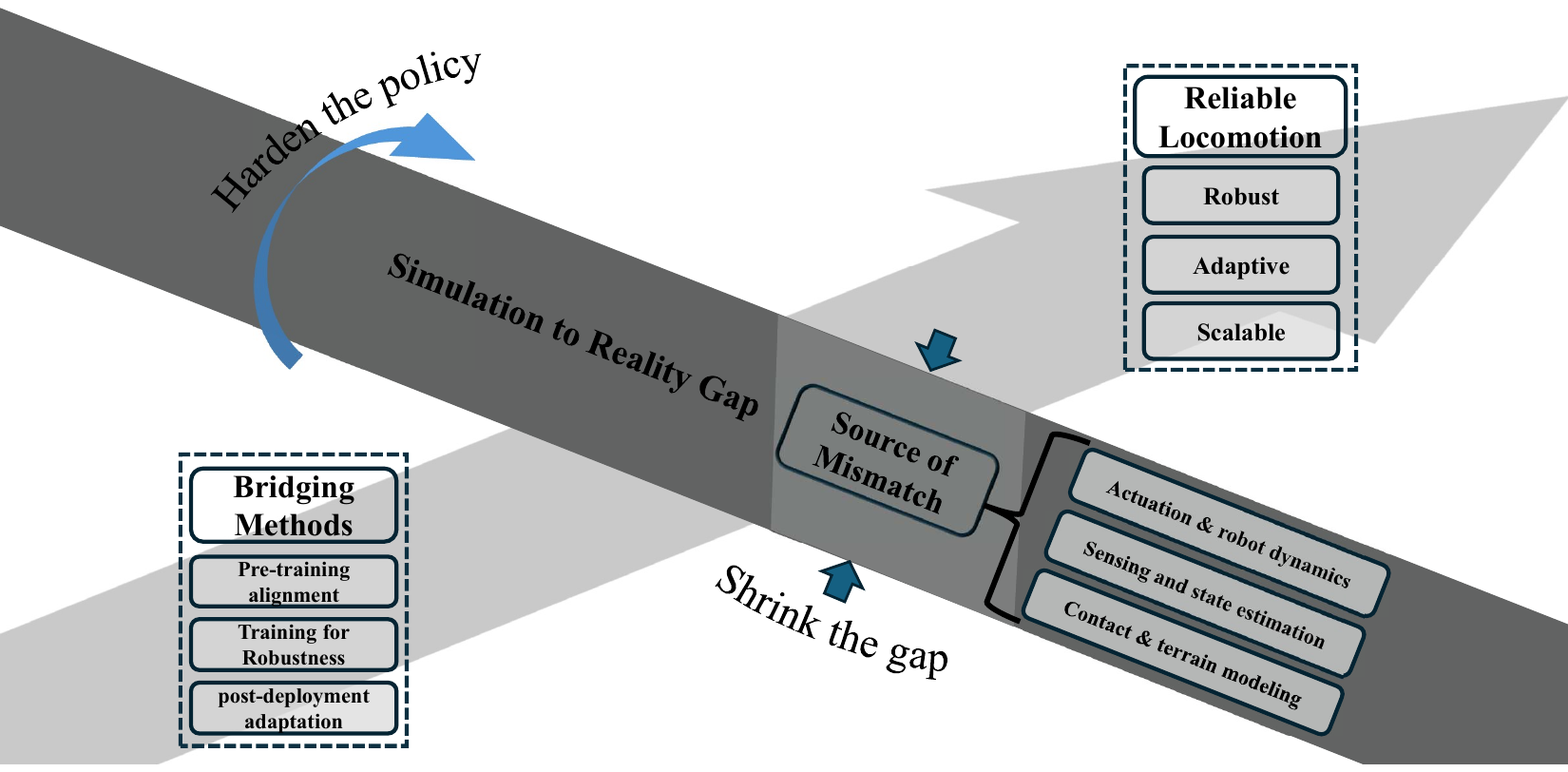}
    \caption{High-level roadmap for sim-to-real bipedal locomotion. Two levers help reduce the transfer gap: (1) shrinking the gap by improving simulator fidelity and system identification, and (2) hardening the policy through domain randomization (DR) and curriculum learning. Residual discrepancies are managed through online adaptation during deployment. Representative sources of mismatch include actuation and robot dynamics, sensing and state estimation, and contact and terrain modeling. The ultimate goal is to achieve robust, adaptive, and scalable bipedal locomotion.}
    \label{fig:sim2real_overall}
\end{figure*}

This section revisits the questions posed in Section~\ref{sec:introduction} regarding sim-to-real transfer. We begin by explaining why training in simulation is generally preferred. We then identify the principal sources of the sim-to-real gap and discuss their implications for improving simulator fidelity. Finally, we outline a pragmatic strategy that balances accuracy and robustness through three complementary components: (i) pre-training alignment, which improves simulator fidelity and encodes nonprivileged information into the policy; (ii) domain randomization (DR) with curriculum learning, which spans plausible uncertainties during training; and (iii) online adaptation, which enables the robot to handle environmental changes at deployment. Figure~\ref{fig:sim2real_overall} summarizes this roadmap.

\subsection{Why Sim-to-Real Training?}

Training locomotion policies directly on physical bipedal robots is prohibitively expensive and unsafe~\cite{2021_Julian_Levine_howtotrainyourrobot_drl}. DRL typically requires tens of millions of interactions to produce stable gaits, which would correspond to months of continuous operation if performed entirely on hardware. Such prolonged trial-and-error accelerates mechanical wear, increases the likelihood of catastrophic failures, and demands continuous human supervision for resets and safety monitoring. On-robot learning is further constrained by partial observability, the difficulty of designing informative reward functions, and the rarity of safety-critical disturbances that are essential for robustness.

Simulation offers a practical alternative. Modern physics engines support massively parallel and faster-than-real-time rollouts, enabling policies to accumulate experience orders of magnitude faster than in the real world. Simulators also allow safe exploration of failure cases, controlled curricula that progressively increase task difficulty, full access to ground-truth states for dense rewards and diagnostics, and reproducible experiments across research groups.

Consequently, nearly all modern locomotion research adopts a sim-to-real workflow: policies are first trained in simulation, then adapted, calibrated, or otherwise transferred to hardware. This paradigm balances data efficiency with safety while motivating the transfer methodologies discussed throughout this chapter.

\subsection{The Curse of Simulation}
\label{sec:cause_sim2real}

To understand why transfer remains challenging, we must examine why even high-fidelity simulators deviate from reality. Despite detailed models of the robot, terrain, and environment, residual discrepancies persist. These mismatches arise primarily from modeling errors in robot dynamics and actuation, interactions with the environment, terrain compliance, and sensing. Additional deviations originate from numerical artifacts introduced by discretization, integration schemes, and contact solvers. The remainder of this subsection details each source and its practical implications.

\paragraph{Robot Dynamics Modeling and Actuation}

The robot’s dynamic model is a primary contributor to the sim-to-real gap. The rigid-body dynamics, expressed in the Lagrangian form in Equation~\ref{eq:rigid_body_dynamics}, highlight the major components where modeling errors can arise:

\begin{equation}
\underbrace{M(q)\ddot{q}}_{\text{Inertia term}}
+ \underbrace{h(q,\dot{q})}_{\text{Coriolis, gravity}}
= \underbrace{S^\top \tau}_{\text{Actuation torques}}
+ \underbrace{J_c(q)^\top \lambda}_{\text{Contact forces}}
+ \underbrace{\tau_{\text{ext}}}_{\text{External disturbance}}
\label{eq:rigid_body_dynamics}
\end{equation}
where \(M(q)\) is the mass matrix, \(h(q,\dot{q})\) aggregates Coriolis, centrifugal, and gravity effects,  \(S^{\top}\tau\) represents actuation torques, and \(J_{c}^{\top}\lambda\) represents contact wrenches.  High-fidelity simulation begins with consistent coordinate frames and accurate inertial parameters for each link, including the mass $m$, center of mass $r_{\text{com}}$, and inertia tensor $I_{\text{com}}$ expressed about the CoM in the link frame.

For bipedal robots, whose long kinematic chains and floating-base dynamics amplify small discrepancies, inaccuracies in these parameters are particularly consequential. Misestimated inertial properties distort the torques required for balance and gait generation, leading to instability and reduced energy efficiency. Therefore, precise identification of rigid-body parameters is often the most effective first step in narrowing the sim-to-real gap.

Beyond rigid-body modeling, the actuation system represents another major source of mismatch. Simulators frequently idealize actuators as perfect torque sources, neglecting real-world nonlinearities that are critical to locomotion. In practice, the actual torque delivered depends on several nonideal effects: (i) Hardware limits, including saturation and slew-rate constraints caused by finite voltage, current, and thermal boundaries~\cite{2018_JieTan_sim2real_locomotion_Minitaur_quadruped}; and
(ii) Electromechanical dynamics, including finite bandwidth and command delays that effectively filter the input signal.

These effects alter both the timing and magnitude of contact forces and are a primary reason that policies appearing stable in simulation often fail on hardware. The underlying motor dynamics are governed by Equation~\ref{eq:motor-dynamics}:

\begin{equation}
\begin{aligned}
\dot{i} &= \tfrac{1}{L}\Big(v - R\,i - k_e\,\omega_m\Big), \
\tau_m  &= k_t\, i, \
P       &= v\,i, \
\end{aligned}
\label{eq:motor-dynamics}
\end{equation}
where the applied voltage \(v\) and back electromotive force (EMF) \(k_{e}\omega_{m}\) determine the current \(i\) and resulting motor torque \(\tau_{m}\). As Equation~\ref{eq:motor-dynamics} illustrates, these dynamics act as a low-pass filter on torque commands, introducing latency and limiting peak responsiveness. Neglecting this filtering effect during simulation often encourages policies that are overly aggressive or poorly timed, which become unstable when deployed on physical robots. Accurate transfer, therefore, requires explicit modeling of actuator limits and response dynamics calibrated from empirical measurements.

\paragraph{Contact and Terrain Modeling}

Beyond robot dynamics, the next major contributor to the sim-to-real gap is environment interaction. For bipedal locomotion, policy stability is highly sensitive to the modeled contact physics. Even small inaccuracies in contact stiffness, friction, or terrain geometry can alter ground-reaction forces (GRFs) and gait timing, destabilizing otherwise functional policies~\cite{2020_ZhaomingXie_iterativeRL_walk_Cassie_biped, 2024_li_ucb_unifiedframework_PPO_IOput_teacherstudentcompare_IOhistory_e2etraining}. Furthermore, real-world surfaces exhibit nonstationary variations in friction and compliance due to wear, moisture, or surface texture—factors that most static simulation models neglect.

Simulators typically employ one of two paradigms: computationally expensive rigid-body models that capture sharp stick–slip transitions, or more stable but less accurate compliant models that approximate contact with spring–damper systems. Both approaches introduce artifacts. Poorly tuned compliant models can exaggerate impact forces or artificially damp energy, while simplified friction models can bias balance-critical Center of Pressure (CoP) estimates. Because realistic deformable terrain simulation is computationally costly, training is often restricted to rigid, planar surfaces, resulting in policies that fail on real-world compliant terrain such as carpet or grass.

\paragraph{Sensing and State Estimation}
In simulation, locomotion policies are typically trained with privileged, noise-free state information, including true joint angles, velocities, and base poses. On hardware, however, these quantities are available only as estimates obtained from encoders, IMUs, and contact sensors through filtering and sensor-fusion pipelines. Such pipelines introduce quantization, bias drift, colored noise, latency, timestamp jitter, and occasional dropouts, rendering the system effectively partially observable. The resulting estimation errors and delays perturb gait phase timing and feedback gains, causing policies trained on perfect state information to become brittle during deployment.

To mitigate these discrepancies, the training setup should mimic the deployed observation interface and timing as closely as possible. Estimator-like inputs should replace ground-truth states, and the statistics of sensing and filtering (noise, bias, latency, dropouts) should be emulated within physically plausible ranges. Policy architectures should integrate information over time, for instance through short observation histories or recurrence, while any privileged variables are reserved for training-only supervision (e.g., critics or auxiliary losses) and excluded from the deployed actor. Maintaining end-to-end consistency between simulated and real observation pipelines—including units, timestamps, normalization, and filter warm-up behavior—substantially improves robustness to sensing imperfection~\cite{2024_Radosavovic_SciRobotics_humanoid_RL}.


\paragraph{Numerical Solvers}

The choice of numerical solver presents a critical trade-off between simulation fidelity, numerical stability, and computational efficiency~\cite{2012_Todorov_MuJoCo_IROS}. Solver approximations directly contribute to the sim-to-real gap by introducing numerical artifacts that can distort or amplify existing modeling errors. These effects manifest primarily in robot dynamics, where integrator type and step size influence stability and energy conservation, and in contact resolution, where the solver determines the realism of impacts and frictional interactions.

Solver design also shapes several system-level properties that are important for reinforcement learning, including determinism across runs, gradient quality during optimization, and scalability on parallel hardware~\cite{2021_Makoviychuk_IsaacGym_arXiv}. While more accurate solvers improve physical fidelity, they also require greater computational resources, necessitating a careful balance between accuracy and throughput for modern DRL frameworks~\cite{2012_Todorov_MuJoCo_IROS}.

\subsection{Pre-training Alignment}
\label{subsec:pretraining-alignment}

Simply increasing simulator fidelity is rarely sufficient to close the sim-to-real gap, since inevitable parameter uncertainties and unmodeled dynamics remain. To address these residual discrepancies, \textbf{pre-training alignment} is introduced as a suite of strategies applied before policy training to systematically improve model accuracy and reduce the burden on the learning algorithm.

\paragraph{Offline System Identification}
\label{subsec:offline-sysid}

System Identification (SI) enhances simulator fidelity by calibrating its physical parameters before policy training, thereby reducing the need for the learned policy to compensate for modeling errors. The procedure is typically formulated as a risk minimization problem. Using data collected from real-world experiments—often designed as ``excitation trajectories'' to expose model inaccuracies—an optimization algorithm searches for the parameters $\theta$ (and, optionally, a system delay $\Delta$) that best reproduce the observed robot behavior. The objective minimizes a loss function that quantifies the discrepancy between simulated and real outputs, such as joint kinematics and contact forces.

\begin{align}
\theta^\star,\Delta^\star
&= \arg\min_{\theta\in\Theta,\;\Delta}\; J(\theta,\Delta) \notag\\
&= \sum_{k=1}^{N_c}\sum_{t=0}^{H_k-1}
   \ell\!\Big(y^{\mathrm{real}}_{t,k},
   \hat y^{\mathrm{sim}}_{t,k}(\theta,\Delta)\Big)
   + \|\theta-\theta_0\|^2_{W_\theta}.
\end{align}

The scope of SI spans inertial, actuation, contact, and sensing parameters, as summarized in Table~\ref{tab:SI_parameters}. A successful SI workflow produces two primary outcomes: a calibrated nominal parameter set $\hat{\theta}$ for a higher-fidelity simulator, and uncertainty estimates that provide principled bounds for DR during policy training.

\begin{table}[htbp]
\caption{System identification parameters in bipedal robot sim-to-real transfer.\label{tab:SI_parameters}}{%
\begin{tabular}{@{}ll@{}}
\toprule
\textbf{Category} & \textbf{Parameters} \\ 
\midrule
Inertial  & Mass; CoM; inertia tensor; link geometry; payload \\ 
\hline
Actuation & Torque constant; gear ratio; friction; backlash; \\
          & compliance; delay; saturation \\ 
\hline
Contact   & Stiffness; damping; friction coefficient; foot sole compliance; \\
          & toe/heel joints; terrain properties \\ 
\hline
Sensing   & Noise; bias; drift; latency; resolution; \\
          & extrinsic calibration; FT calibration \\ 
\hline
\end{tabular}}{}
\end{table}

In practice, the SI process often combines analytical measurements with whole-body TO. The workflow typically begins with inertial calibration to correct deviations in the CAD model, a procedure that may be conducted both before and after initial policy training to incorporate task-relevant data~\cite{2019_Yuwenhao_sim2real_locomotion_DarwinOP2_humanoid}. More critically, hardware-specific properties must be identified. For instance, reproducing realistic ground contact for the Cassie biped required experiments to determine the unique stiffness and damping characteristics of its compliant feet \cite{2020_JonahSiekmann_memoryRL_walk_Cassie_biped}. Similarly, characterizing actuation dynamics—such as motor constants and friction—is essential for matching the robot’s torque response and achieving stable gaits~\cite{2020_ZhaomingXie_iterativeRL_walk_Cassie_biped}.

\paragraph{Residual Dynamics Learning}
Residual discrepancies caused by unmodeled effects such as nonlinear friction, contact dynamics, or actuator thermal variations inevitably remain after initial system identification. Residual Dynamics Learning provides a complementary, model-centric approach to address this remaining gap. Instead of re-identifying all of the simulator’s base parameters, this method focuses on learning a direct correction to the simulator’s dynamics model to compensate for residual mismatches:

\begin{equation}
f_{\text{real}}(s_t, a_t) \approx f_{\text{sim}}(s_t, a_t) + g_{\phi}(s_t, a_t).
\label{eq:residual_model}
\end{equation}

where $s_t$ and $a_t$ denote the state and action at time $t$. The function $g_{\phi}(s_t, a_t)$, typically parameterized by a neural network with parameters $\phi$, outputs a predicted correction $\Delta s_t$ to the simulator’s next-state prediction.

For example, \cite{2023_Nitish_BALLU_physicslearning} proposed a Residual Physics Learning framework that augments traditional system identification with a learned residual dynamics term. In their approach, the nominal simulator captures coarse rigid-body and buoyancy effects, while a neural residual model explains unmodeled forces and contacts. In contrast, a policy-centric formulation learns a residual directly in the action space. Instead of correcting the dynamics model, methods such as ASAP learn a delta-action model that directly adjusts the output of a simulation-trained policy \cite{2025_he_ASAP}.

\subsection{Training for Robustness}
While pre-training alignment seeks to reduce the sim-to-real gap by improving the simulator or performing initial policy corrections, an alternative and complementary philosophy is to train a policy that is inherently robust to this gap from the outset. The primary objective of these in-training strategies is to enable zero-shot sim-to-real transfer, where the learned policy can be deployed directly to physical hardware without any real-world fine-tuning. This section discusses two major paradigms that address these challenges: DR and teacher–student learning.

\paragraph{Domain Randomization}
\label{subsec:domain_randomization_curriculum}

To account for the inevitable discrepancies between simulation and reality, DR trains a policy across a wide distribution of simulated environments rather than a single deterministic one. The core idea is that if the randomized distribution is sufficiently broad to include the real world’s dynamics as one of its instances, the learned policy should transfer successfully without fine-tuning. This approach promotes robustness to variations but can also yield more conservative behaviors and requires significantly more training data.

In practice, DR involves randomizing a set of parameters at the start of each training episode. These parameters typically fall into two categories:

\begin{enumerate}
    \item Dynamics parameters: Physical properties such as link masses and inertias, joint friction, motor strength, and actuator delays are varied within predefined ranges.

    \item Environmental conditions: Factors such as random external forces, variations in initial states, sensor noise, latency, and ground friction are randomized to simulate real-world uncertainty.

\end{enumerate}

The choice of randomization ranges is critical. Ranges that are too narrow may fail to cover the sim-to-real gap, while overly broad ranges can make the control problem intractable and lead to overly conservative or unstable policies.

DR has proven highly effective for sim-to-real transfer in legged robots. For example, controllers for the ANYmal quadruped have successfully employed a loose calibration plus DR strategy, in which core parameters are identified while difficult-to-model properties such as friction are heavily randomized \cite{2020_JoonhoLee_memoryRL_locomotion_ANYmal_quadruped}. Similarly, the DeepWalk policy learned to stabilize the NimbRo-OP2X biped by training with random external pushes, allowing the real robot to withstand disturbances \cite{2021_DiegoRodriguez_deepRL_locomotion_NimbRoOP2X_humanoid}. 

To mitigate the training difficulty associated with wide randomization, Curriculum Learning (CL) is often used as a complementary strategy. Rather than exposing the policy to the full range of variations from the start, the agent is trained on progressively harder tasks. This can involve gradually widening the randomization ranges or increasing the target velocity, as demonstrated in the ALLSTEPS framework for the Cassie biped \cite{2020_zhaoming_drl_steppingstones_PPOwithactorcritic_referencefree_simulation}.

\paragraph{Teacher-student Paradigm}
Beyond physical uncertainty, another central challenge in sim-to-real transfer is perceptual uncertainty. An agent in simulation has access to perfect, global, privileged information, such as the precise CoM velocity, ground friction coefficient, or exact contact forces. A real robot, however, must rely on noisy, delayed, and incomplete onboard sensors for state estimation. Training a policy directly across this large information gap often leads to failure.

The teacher–student paradigm is a knowledge distillation technique specifically designed to address this issue. The core idea is to train two policies in simulation:  
(i) the teacher policy, $\pi_{\text{teacher}}$, which has access to all privileged information and can therefore learn near-optimal, expert-level behavior; and  
(ii) the student policy, $\pi_{\text{student}}$, which is also trained in simulation but is limited to the non-privileged sensory inputs that a real robot would have, such as simulated IMU readings, joint encoder angles, and camera images.

During training, the student does not learn directly from environmental interaction. Instead, it learns to reproduce the teacher’s actions through supervised learning, minimizing the discrepancy between its predicted actions and those generated by the teacher. The student thus learns a mapping function that uses only partial and noisy observations to approximate the expert’s decision-making process.

Ultimately, the student policy is deployed on the real robot, as it has been trained to perform expert-level behaviors under realistic perceptual constraints.

\subsection{Post-deployment Adaptation}
\label{subsec:adapation_residual}

Even with high-fidelity simulation, offline system identification, and DR, learned policies are typically robust only within the distribution of conditions encountered during training. They perform well in environments that closely resemble the training domain, but real-world deployment inevitably introduces gradual drifts such as changes in ground friction, actuator heating, payload variations, or mechanical wear. These factors create residual mismatches and can degrade performance over time.

Online adaptation addresses this issue by adjusting policy behavior during execution using real-time feedback from robot–environment interactions. Approaches in this category can generally be divided into two main families: system identification–based adaptation, which reduces residual mismatches by either explicitly estimating time-varying physical parameters (for example, inertia, actuation, contact, and sensing) or implicitly inferring latent context variables from short observation–action histories; and adaptive policy learning, which directly adjusts the control policy through online fine-tuning, residual reinforcement learning, or other feedback-driven mechanisms.

\paragraph{Explicit Online System Identification}

Online system identification can be broadly divided into two families: explicit approaches, which estimate physically meaningful parameters, and implicit approaches, which infer latent context representations from short interaction histories. This section introduces both perspectives and outlines their mechanisms, mathematical formulations, and representative applications in bipedal locomotion.

Explicit approaches directly estimate time-varying physical parameters such as inertia, CoM, actuation constants, contact stiffness, and sensor biases, allowing the underlying model or controller to remain calibrated during deployment. Common techniques include recursive least squares (RLS), disturbance observers, and extended or unscented Kalman filters (EKF/UKF). Formally, the estimation process can be expressed as
\[
\hat{\phi}_t = E_\psi(h_t), \quad a_t \sim \pi_\theta(s_t; \hat{\phi}_t),
\]
where $\hat{\phi}_t$ denotes the online estimate of the physical parameters, $h_t$ is a short trajectory segment, and $\pi_\theta$ is the policy conditioned on the updated parameters.

Representative examples include Kalman filter--based estimation of  inertial and contact properties in bipedal locomotion~\cite{2019_JeminHwangbo_sim2real_locomotion_ANYmal_quadruped},  as well as recent work on physically consistent EKF adaptation for humanoid loco-manipulation~\cite{Foster2024_PhysicallyConsistentEKF}.

\paragraph{Implicit Online System Identification} 

Implicit approaches do not attempt to recover the physical parameters \(\phi\) explicitly. Instead, they infer a compact latent variable from recent observation–action histories and condition the policy on it to achieve online adaptation. Formally,
\[
z_t = g_\psi(h_t), \qquad a_t \sim \pi_\theta(o_t, z_t), \qquad
h_t = (o_{t-H:t},\, a_{t-H:t-1}).
\]
Compared with explicit online system identification, which estimates \(\phi\) and updates controller gains, the implicit approach amortizes system identification into the encoder \(g_\psi\) and emphasizes rapid, task-relevant adaptation.

A useful taxonomy distinguishes three common ways to obtain the context variable \(z_t\) from observation and action histories.  
(i) Recurrent state update: maintain an online latent state \(z_t = f_\psi(z_{t-1}, o_t, a_{t-1})\), where \(f_\psi\) is implemented using a gated recurrent unit, long short-term memory network, or a learned state-space update.  
(ii) Sliding-window encoding: construct \(h_t = (o_{t-H:t}, a_{t-H:t-1})\) and map it through a temporal encoder \(z_t = g_\psi(h_t)\); common choices include temporal convolutions or causal attention encoders.  
(iii) Latent filtering: treat \(z_t\) as a belief state with learned prior and posterior distributions, \(p_\psi(z_t \mid z_{t-1}, a_{t-1})\) and \(q_\psi(z_t \mid z_{t-1}, a_{t-1}, o_t)\), updated in a Kalman-like manner and trained using predictive or variational objectives.  
In practice, \(z_t\) is often smoothed using exponential moving averages and temporally regularized; resets are triggered when prediction residuals or \(\lVert \Delta z_t \rVert\) exceed predefined thresholds.

Memory-based approaches with recurrent policies provide strong evidence that encoding observation and action histories enables adaptation under partial observability. For example, an end-to-end LSTM policy was transferred to the biped Cassie in zero-shot fashion~\cite{2021_siekmann_sim2real_nonreference_perodicreward_DRL_e2e_LSTM_PPO_cassie}. Rapid Motor Adaptation (RMA) combines a base locomotion policy with an adaptation module that infers a low-dimensional context from recent histories, allowing quadruped robots to adapt in real time across diverse terrains~\cite{2021_AshishKumar_RMA_loc_UnitreeA1_quadruped}. Building on this idea, A-RMA extends the method to bipedal locomotion, where the context-conditioned policy enables fast adaptation to changing loads and terrains~\cite{2022_ARMA_UCB_onlineadaption_bipedal_cassie}. More recently, transformer-based in-context adaptation has scaled such history-conditioned mechanisms to humanoid robots~\cite{2024_Radosavovic_SciRobotics_humanoid_RL}.

\section{Discussion}
\label{sec:discussion}
The sim-to-real approaches reviewed in this chapter are not isolated solutions but interconnected components of a unified strategy, organized by their application stage from pre-training alignment to post-deployment adaptation. The success of this strategy rests on two complementary philosophies: reducing the sim-to-real gap through higher model fidelity and making the policy resilient to the errors that inevitably remain. This discussion synthesizes these ideas, outlining how these tools work together in practice, where each method helps, where it falls short, and what is needed to advance the field.

\subsection{Sim-to-Real Methods Limits}
While each sim-to-real technique provides a powerful solution to a specific aspect of the sim-to-real gap, a critical analysis reveals that no single method serves as a universal remedy. In practice, relying exclusively on one approach exposes a distinct set of limitations. A clear understanding of these individual shortcomings is therefore essential for navigating the complex landscape of sim-to-real transfer.

Challenges already arise at the pre-training alignment stage. System identification is essential but labor-intensive, and it struggles to capture complex, unmodeled dynamics or slow nonstationary drift due to wear or recalibration. Residual learning is often introduced to compensate, yet it introduces a different trade-off: dependence on real-world data collection, which is costly and safety-constrained, and corrections that are narrow in scope, often failing to generalize beyond the training envelope or to new tasks.

Similarly, training-for-robustness strategies bring their own compromises. DR, the cornerstone of zero-shot transfer, can produce overly conservative policies that sacrifice peak performance for generality. Moreover, selecting effective randomization ranges remains a poorly scaled, heuristic-driven process. The teacher–student paradigm, while effective for addressing perceptual uncertainty, is limited by the teacher’s maximum achievable performance and does not, by itself, resolve mismatches in physical dynamics.

Post-deployment adaptation can handle long-term changes but introduces the classic stability–plasticity dilemma: the agent must adapt without catastrophically forgetting prior knowledge. This adds complexity and raises safety concerns, since a continually adapting policy is inherently non-stationary and difficult to verify. In short, every paradigm imposes trade-offs—whether in data dependency, performance conservatism, or algorithmic complexity. The most practical solutions arise not from selecting a single method but from intelligently integrating multiple approaches.

\subsection{From Theory to Practice}
The most effective sim-to-real solutions emerge from the synergistic combination of complementary methods rather than reliance on any single technique. A dominant strategy is to pair a high-fidelity baseline with robust training. This typically involves first applying offline system identification or residual dynamics learning to construct a more accurate simulator, which then serves as the foundation for training more robust policies~\cite{2023_Nitish_BALLU_physicslearning}. This initial alignment allows for narrower and more targeted randomization, mitigating the policy conservatism often associated with DR.

A second, complementary paradigm produces policies that are robust to both perceptual limitations and dynamic environmental variations. This is achieved by combining the teacher–student framework, which hardens the policy against sensor noise, with implicit online system identification~\cite{2023_vanmarum_visionDRL_studentteacher_irregularterrain_PPO_periodicrewardfunction,2024_Radosavovic_SciRobotics_humanoid_RL}. In this setup, an encoder uses the robot’s recent history to infer the current real-world dynamics, allowing the policy to adapt its behavior online. This combination of robust perception and continuous adaptation forms the foundation of several state-of-the-art systems~\cite{2024_li_ucb_unifiedframework_PPO_IOput_teacherstudentcompare_IOhistory_e2etraining}.

\subsection{Future Outlook}
The trajectory of sim-to-real development can be viewed as an evolutionary progression through three stages. The field has advanced from early Stage~1, Simulation plus Fine-tuning, which was hindered by the inefficiency and risks of real-world learning, to the current dominant paradigm, Stage~2, Zero-Shot Transfer. Although techniques such as DR have enabled much of today’s agile locomotion, the resulting policies are often conservative and brittle when faced with out-of-distribution conditions.

The next frontier is Stage~3, Real-World Mastery, in which robots progress beyond transferring a fixed skill set to continuously adapting and improving in open-world environments. This stage envisions agents that achieve genuine competence by integrating extensive simulated experience with rapid online learning, enabling them to handle novel situations long after their initial deployment.

The key challenge lies in transitioning from the static paradigm of Stage~2 to the dynamic paradigm of Stage~3. This shift requires moving beyond one-way transfer toward a continuous learning loop, supported by concurrent advances in three areas. It will require better models, evolving from static simulators into adaptive digital twins that are continuously updated with real-world data. It will require better policies, advancing from task-specific optimization toward generalization through improved algorithms and architectures. Above all, it will demand better adaptation, requiring online learning algorithms that are not only fast and data-efficient but also provably safe, establishing the foundation for true lifelong robotic learning.

\section{Conclusions}\label{sec:conclusion}

Sim-to-real for bipedal locomotion is not a bag of tricks but a principled approach to modeling, learning, and adaptation. Whether this is implemented within an end-to-end framework or a multi-layered hierarchical scheme, the core challenges of bridging the sim-to-real gap remain the same. The most effective way to navigate these challenges is to focus on two primary lines of effort, reinforced by a continuous adaptation loop. The first line of effort targets the Model: shrinking the sim-to-real gap by raising simulator fidelity through careful identification. The second targets the policy: hardening it against residual errors by training for variability. The adaptation loop ensures that once deployed, the robot can continuously read the world and adjust to keep its behavior robust and safe.

In practice, this means focusing on the model first by calibrating a verifiable nominal model. The focus then shifts to the policy, training with measurement-driven DR and mirroring the real observation pipeline. Finally, the adaptation loop is closed by equipping the controller with mechanisms for online inference and adjustment. Each of these efforts carries inherent trade-offs: identification is laborious, broad randomization can yield conservative policies, and adaptation complicates verification. A disciplined blend is therefore more reliable than any single method.

The outlook for the field is one of increasing integration and capability across all three fronts. Progress will depend on advances in simulation, policy design, and adaptation. Future simulators will evolve from static tools into continuously updated digital twins or generative world models that learn directly from real-world data. Policies will become more general, drawing on large-scale foundation models to move beyond task-specific training toward versatile, goal-directed competence. Adaptation mechanisms will grow more powerful, enabling online learning that is faster, safer, and more data-efficient. The guiding principle remains clear: begin with physics you can verify, train for the variability you expect, and empower the robot to learn as it walks.


\bibliographystyle{IEEEtran}
\bibliography{Wiley}


\end{document}